\documentclass{article}

\PassOptionsToPackage{numbers, compress, sort}{natbib}

\usepackage[final]{neurips_2019}

\usepackage[utf8]{inputenc} %
\usepackage[T1]{fontenc}    %
\usepackage{hyperref}       %
\usepackage{url}            %
\usepackage{booktabs}       %
\usepackage{amsfonts}       %
\usepackage{nicefrac}       %
\usepackage{microtype}      %

\usepackage{macro}

\usepackage{authblk}
\usepackage{microtype}
\usepackage{graphicx}
\usepackage{subfigure}
\usepackage{amsmath}
\usepackage{multirow}
\usepackage{bbm}
\usepackage{caption} %

\title{\systemx: A Programming Model for Residual Learning in Critical Data Slices }

\author[ ]{\textbf{
  Vincent S.~Chen, 
  Sen Wu, 
  Zhenzhen Weng, 
  Alexander Ratner, 
  Christopher R\'e
  }\vspace{-0.2cm}}
\affil[ ]{Stanford University}
\affil[ ]{ \texttt{
  vincentsc@cs.stanford.edu, 
  senwu@stanford.edu, 
  zzweng@stanford.edu, 
  ajratner@stanford.edu,
  chrismre@cs.stanford.edu
  }
}

\begin{document}

\maketitle

\begin{abstract}
In real-world machine learning applications, data subsets correspond
to especially critical outcomes: vulnerable cyclist detections are
safety-critical in an autonomous driving task, and ``question''
sentences might be important to a dialogue agent's language
understanding for product purposes.  While machine learning models can
achieve high quality performance on coarse-grained metrics like F1-score and
overall accuracy, they may underperform on critical subsets---we
define these as \textit{slices}, the key abstraction in our approach.
To address slice-level performance, practitioners often train separate
``expert'' models on slice subsets or use multi-task hard parameter sharing.  We
propose \systemx, a new programming model in which the \textit{slicing
function (SF)}, a programming interface, specifies critical data subsets for which the model should commit additional capacity. Any model can leverage SFs
to learn \textit{slice expert representations}, which are combined with an attention
mechanism to make \textit{slice-aware} predictions.  We show that our approach
maintains a parameter-efficient representation while improving over baselines
by up to \num{19.0 F1} on slices and \num{4.6 F1} overall on datasets spanning
language understanding (e.g. SuperGLUE), computer vision, and production-scale
industrial systems.

\end{abstract}

\section{Introduction}
\label{sec:intro}

In real-world applications, some model outcomes are more important
than others: for example, a data subset might correspond to safety-critical but
rare scenarios in an autonomous driving setting (e.g. detecting
cyclists or trolley cars~\cite{karpathy2018software}) or critical but
lower-frequency healthcare demographics (e.g. bone X-rays associated with degenerative joint disease~\cite{oakden2019hidden}). Traditional machine learning systems optimize for overall
quality, which may be too coarse-grained; models that achieve high
overall performance might produce unacceptable failure rates
on \textit{slices} of the data. In many production settings, the key challenge
is to maintain overall model quality while improving slice-specific metrics.

\begin{figure}[t]
  \begin{center}
    \includegraphics[width=1.0\textwidth]{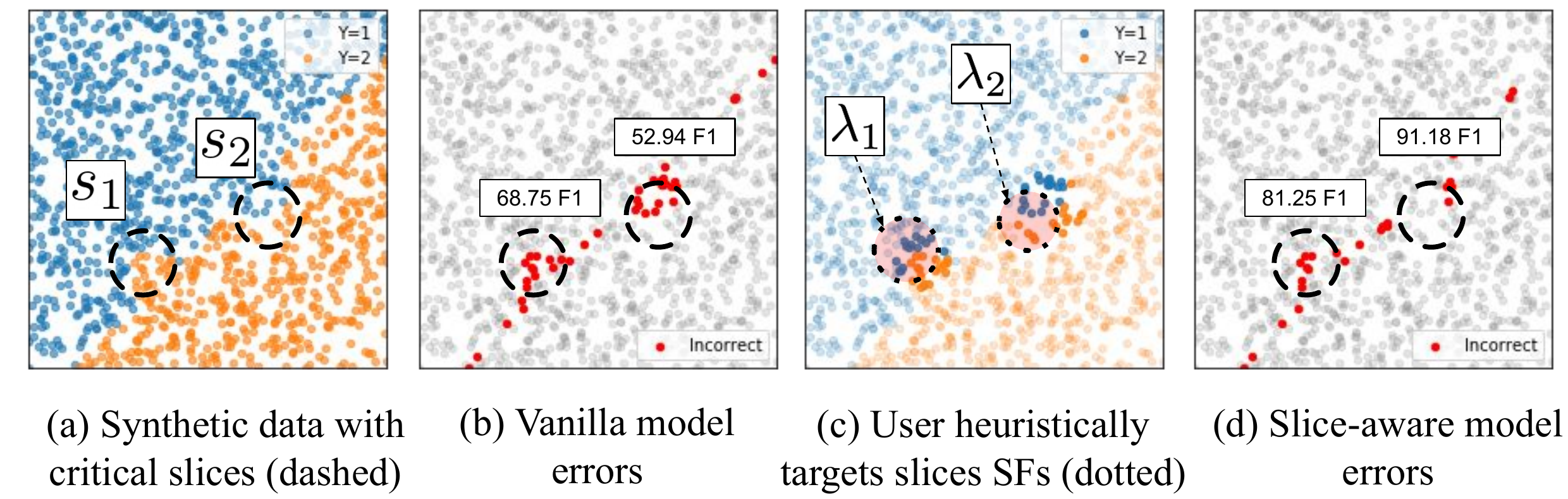}
    \caption{
      \textbf{\systemx via synthetically generated data:}
      (a) The data distribution contains critical slices, $s_1$, $s_2$, that represent a small proportion of the dataset.
      (b) A vanilla neural network correctly learns the general, linear decision boundary but fails to learn the perturbed slice boundary.
      (c) A user writes \textit{slicing functions} (SFs), $\lambda_1$, $\lambda_2$, to heuristically target critical subsets.
      (d) The model commits additional capacity to learn slice expert representations.
      Upon reweighting slice expert representations, the slice-aware model learns to classify the fine-grained slices with higher F1 score.
    }
    \label{fig:synthetics_overview}
  \end{center}
 	\vspace{-0.1in}
\end{figure}

To formalize this challenge, we introduce the notion of
\textit{slices}: application-critical data subsets, specified programmatically by machine learning practitioners, for which  we would like to improve model performance.
This leads to three technical challenges:
\begin{itemize}
\item \textbf{Coping with Noise:} Defining slices precisely can be
challenging. While engineers often have a clear intuition of a
slice, typically as a result of an error analysis,
translating that intuition into a machine-understandable description
can be a challenging problem, e.g., {\it ``the slice of data that
contains a yellow light at dusk.''} As a result, any method must be
able to cope with imperfect, overlapping definitions of data
slices, as specified by noisy or \textit{weak supervision}.

\item \textbf{Stable Improvement of the Model:}
Given a description of a set of slices, we want to improve the
prediction quality on each of the slices without hurting overall
model performance. Often, these goals are in tension: in many baseline
approaches, steps to improve the slice-specific model performance would
degrade the overall model performance, and vice-versa.

\item \textbf{Scalability:} There may be many slices. Indeed, in industrial
deployments of slice-based approaches, hundreds of
slices are commonly introduced by engineers~\cite{re2019overton}---any approach to \systemx must be judicious with adding parameters as the
number of slices grow.
\end{itemize}

To improve \textit{fine-grained}, i.e. slice-specific, performance, an
intuitive solution is to create a separate model for each slice. To
produce a single prediction at test time, one often trains a \textit{mixture of
experts} model (MoE)~\cite{jacobs1991adaptive}. However, with the growing size
of ML models, MoE is often untenable due to runtime performance, as it
could require training and deploying hundreds of large models---one for
each slice.  Another strategy draws from multi-task learning (MTL), in
which slice-specific \textit{task heads} are learned with
hard-parameter sharing~\cite{caruana1997multitask}. This approach is
computationally efficient but may not effectively share training data
across slices, leading to suboptimal performance. Moreover, in MTL,
tasks are distinct, while in \systemx, a single \textit{base task} is
refined by related slice tasks.

We propose a novel programming model, called \systemx, in which
practitioners provide slicing functions (SFs), a programming
abstraction for heuristically targeting data subsets of interest.
SFs coarsely map input data to slice indicators, which specify
data subsets for which we should allocate additional model
capacity. To improve slice-level performance, we introduce {\em slice-residual-attention modules} (SRAMs) that explicitly model 
\textit{residuals} between slice-level and the overall task
predictions. SRAMs are agnostic to the
architecture of any neural network model that they are added to---which
we refer to as the \textit{backbone} model---and we
demonstrate our approach on state-of-the-art text and image
models. Using shared backbone parameters, our model initializes 
slice ``expert'' representations, which are associated with learning
slice-membership indicators and class predictors for examples in a
particular slice. Then, slice indicators and prediction confidences
are used in an \textit{attention-mechanism} to reweight and combine
each slice expert representation based on learned residuals from the base representation.
This produces a \textit{slice-aware} featurization of the data, which can be used to make a final prediction.

Our work fits into an emerging class of programming models that sit on
top of deep learning systems~\cite{ratner2019role,karpathy2018software}. 
We are the first to introduce and formalize \systemx, a key programming
abstraction for improving ML models in real-world applications subject to
slice-specific performance objectives. Using an independent
error analysis for the recent GLUE natural language understanding benchmark
tasks \cite{wang2018glue}, by simply encoding the identified error categories
as slices in our framework, we show that we can improve the quality of
state-of-the-art models by up to \num{4.6} F1 points, and we observe slice-specific
improvements of up to \num{19.0} points. We also evaluate our system on
autonomous vehicle data and show improvements up to \num{15.6 F1 points} on
context-dependent slices (e.g., presence of bus, traffic light) and
\num{2.3 F1 points} overall. Anecdotally, when
deployed in production systems~\cite{re2019overton}, \systemx provides a
practical programming model with improvements of up to \num{40 F1 points} in
critical test-time slices. On the SuperGlue benchmark~\cite{wang2019superglue},
this procedure accounts for a \num{2.7} improvement in aggregate benchmark score using
the same architecture as previous state-of-the-art submissions.
In addition to the proposal of SRAMs, we perform an in-depth analysis to
explain the mechanisms by which SRAMs improve quality. We validate the
efficacy of quality and noise estimation in SRAMs and compare to weak
supervision frameworks~\cite{ratner2019role} that estimate the quality of
supervision sources to improve overall model accuracy. We show that by using
SRAMs, we are able to produce accurate quality estimates, which leads to higher
downstream performance on such tasks by an average of \num{1.1 overall F1 points}.

\section{Related Work}
\label{sec:related-work}
Our work draws inspiration from three main areas: mixture of experts,
multi-task learning, and weak
supervision. Jacobs et. al~\cite{jacobs1991adaptive} proposed a technique
called \textbf{mixture of experts} that divides the data space into
different homogeneous regions, learns the regions of data
separately, and then combines results with a single gating
network~\cite{sigaud2015gated}.  This work is a generalization of
popular ensemble methods, which have been shown to improve predictive
power by reducing overfitting, avoiding local optima, and combining
representations to achieve optimal
hypotheses~\cite{sagi2018ensemble}. We were motivated in part by
reducing the runtime cost and parameter count for such models.

\textbf{Multi-task learning}
(MTL) models provide the flexibility of \textit{modular}
learning---specific task heads, layers, and representations can be
changed in an application-specific, ad hoc manner.  Furthermore, MTL
models benefit from the computational efficiency and regularization
afforded by hard parameter sharing~\cite{caruana1997multitask}.  There
are often also performance gains seen from adding auxiliary tasks to
improve representation learning objectives~\cite{cheng2015open,rei2017semi}.
While our approach draws high-level inspiration from MTL, we highlight key
differences: whereas tasks are disjoint in MTL, slice tasks are
formulated as \textit{micro-tasks} that are direct extensions of a base
task---they are designed specifically to learn deviations from the
base-task representation.  In particular, sharing information, as seen
in cross-stitch networks~\cite{misra2016cross}, requires $\Omega(n^2)$
weights across $n$ local tasks; our formulation only requires
attention over $O(n)$ weights, as slice tasks operate on
the \textit{same} base task.  For example, practitioners might specify
yellow lights and night-time images as important slices; the model
learns a series of micro-tasks---based solely on the data
specification---to inform how its approach for the base task, object
detection, should change in these settings.  As a result, slice tasks are
not fixed ahead of time by an MTL specification; instead, these micro-task
boundaries are learned dynamically from corresponding data
subsets. This style of information sharing sits adjacent to cross-task
knowledge literature in recent MTL models~\cite{yang2016deep,ruder2017overview}, and we were inspired by
these methods.

\textbf{Weak supervision} has been viewed as a new way to incorporate data
of varying accuracy sources, including domain experts, crowd sourcing,
data augmentations, and external knowledge bases
\cite{ratner2016data,bach2018snorkel,ratner2018snorkel,mintz2009distant,Berend:2014:CWM:2969033.2969211,Dalvi:2013:ACB:2488388.2488414,NELL-aaai15,blum1998combining,mann2010generalized}.
We take inspiration from {\em labeling functions}
\cite{ratner2016data} in weak supervision as a programming paradigm, which has
seen success in industrial deployments~\cite{bach2018snorkel}.
In existing weak supervision literature, a key challenge is
to assess the accuracy of a training data point, which is a function
of supervision sources. In this work, we model this
accuracy using learned representations of user-defined slices---this leads to
higher overall quality.

Weak supervision and multitask learning can be viewed as
orthogonal to slicing: we have observed them used alongside \systemx
in academic projects and industrial deployments~\cite{re2019overton}.

\section{Slice-based Learning} %
\label{sec:data-slicing}
\begin{figure}[t]
  \begin{center}
    \includegraphics[width=1.0\textwidth]{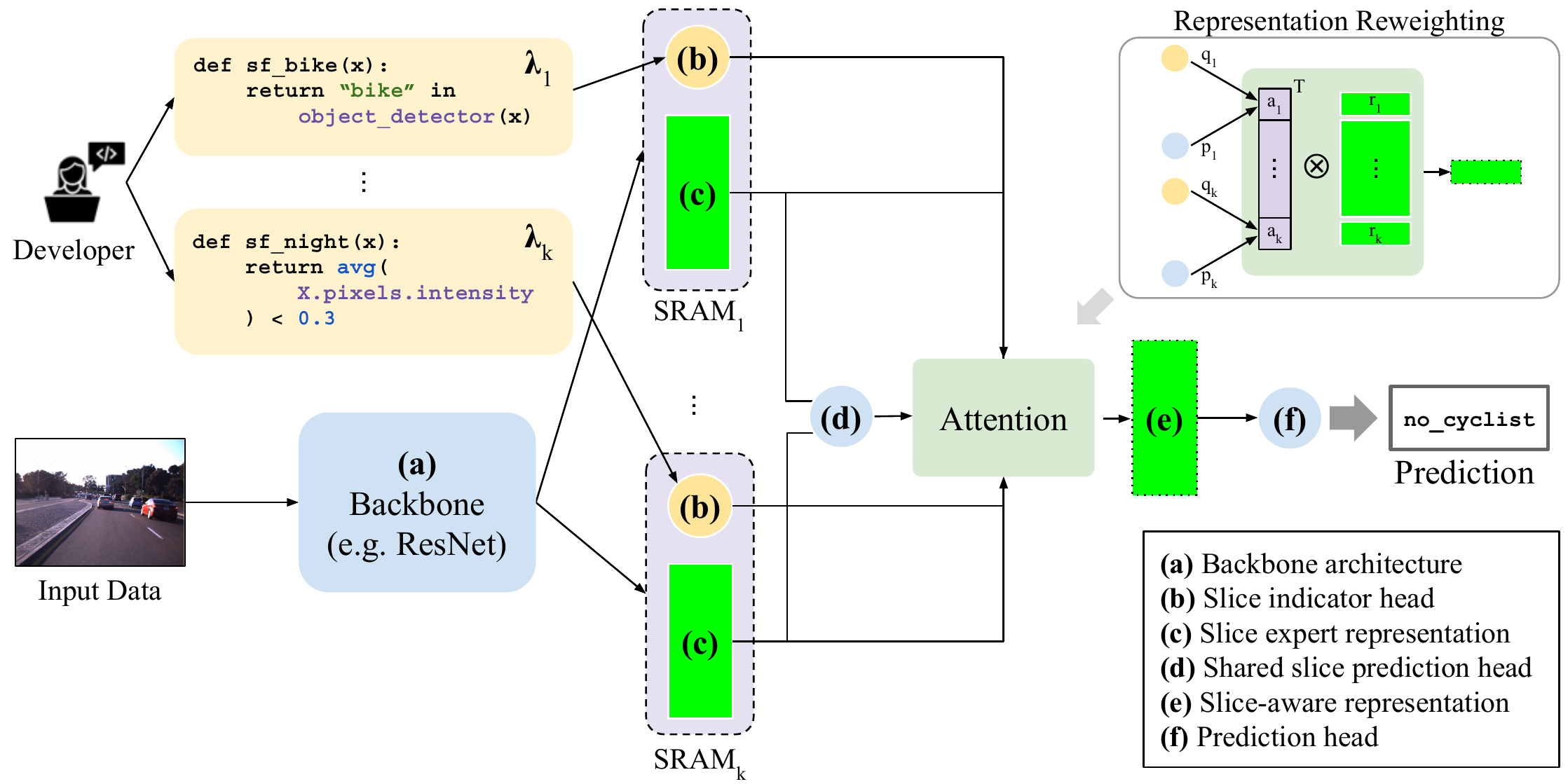}
    \caption{
      \textbf{Model Architecture}:
      A developer writes SFs ($\lambda_{i=1,\dots,k}$) over 
      input data and specifies any \textbf{(a) backbone} architecture (e.g.
      ResNet~\cite{he2016deep}, BERT~\cite{devlin2018bert}) as a feature
      extractor. These features are shared parameters for $k$ slice-residual
      attention modules (SRAMs); each learns a \textbf{(b) slice indicator head} which outputs a prediction, $q_i$, indicating which slice the example belongs to, as supervised by $\lambda_i$.
      SRAMs also learn a \textbf{(c) slice expert representation}, trained only on examples belonging to the slice using a
      \textbf{(d) shared slice prediction head}, which makes predictions, $p_i$, on the original task schema and is supervised by the masked ground truth labels for the corresponding slice.
      An attention mechanism, $a$, reweights these
      representations, $r_i$, into a combined, \textbf{(e) slice-aware 
      representation}. A final \textbf{(f) prediction head} makes
      model predictions based on this slice-aware representation.
    }
    \label{fig:architecture}
  \end{center}
     \vspace{-0.1in}
\end{figure}

We propose \textit{\systemx} as a programming model for training machine
learning models where users specify important data subsets to improve model
performance. We describe the core technical challenges that lead to our notion
of {\em slice-residual-attention modules} (SRAMs).

\subsection{Problem statement}
To formalize the key challenges of slice-based learning, we introduce
some basic terminology. In our \textit{base task}, we use a supervised input,
$(x \in \mathcal{X}, y \in \mathcal{Y})$, where the goal is to learn
according to a standard loss function. In addition, the
user provides a set of $k$ functions called {\em slicing functions (SFs)},
$\{\lambda_1,\dots,\lambda_k\}$, in which $\lambda_i : {\cal X} \to
\{0,1\}$. These SFs are not assumed to be perfectly accurate;
for example, SFs may be based on noisy or \textit{weak}
supervision sources in functional form~\cite{ratner2016data}.
SFs can come from domain-specific heuristics, distant supervision
sources, or other off-the-shelf models, as seen in 
Figure~\ref{fig:architecture}.
Ultimately, the model's goal is to improve (or avoid damaging) the overall
accuracy on the base task while improving the model on the specified slices.

Formally, each of $k$ slices, denoted $s_{i=1,\dots,k}$, is an unobserved,
indicator random variable, and each user-specified SF,
$\lambda_{i=1,\dots,k}$ is a corresponding, noisy specification.
Given an input tuple $(\mathcal{X}, \mathcal{Y}, \{\lambda_i\}_{i=1,\ldots,k})$ consisting of a dataset $(\mathcal{X}, \mathcal{Y})$, and $k$ different user-defined SFs $\lambda_i$, our goal is to learn a model $f_{\hat{w}}(\cdot)$---i.e. estimate model parameters $\hat{w}$---that predicts $P(Y|\{s_i\}_{i=1,\ldots,k},\mathcal{X})$ with high slice-specific accuracies without substantially degrading overall accuracy.

\begin{example}
\label{ex:yellow_light}
A developer notices that their self-driving car is not detecting
cyclists at night. Upon error analysis, they diagnose that their
state-of-the-art object detection model, trained on an automobile
detection dataset $(\mathcal{X}, \mathcal{Y})$ of images, is indeed
underperforming on \textit{night} and \textit{cyclist}
slices. They write two SFs: $\lambda_1$ to classify night vs. day, based
on pixel intensity; and $\lambda_2$ to detect bicycles, which calls a
pretrained object detector for a bicycle (with or without a rider). 
Given these SFs, the developer leverages \systemx to improve model performance on safety-critical subsets.
\end{example}

Our problem setup makes a key assumption: SFs may be \textit{non-servable}
during test-time---i.e, during inference, an SF may be unavailable because it is too expensive to compute or relies on private metadata
~\cite{bach2019snorkel}. In Example~\ref{ex:yellow_light}, the potentially
expensive cyclist detection algorithm is non-servable at runtime. When our model is served at inference, \textit{SFs are not necessary}, and we can rely on the model's \textit{learned} indicators.

\begin{table}[t]
  \begin{minipage}{.45\linewidth}
    \tiny
    \begin{tabular}{lccccc}\\
      \toprule
      \multirow{2}{*}{\parbox{2.5cm}{\textbf{Reweighting Mechanism}}} & \multicolumn{5}{c}{Performance (F1 score)} \\
      & Overall & $s_1$ & $s_2$& $s_3$ & $s_4$ \\
      \midrule
      \textsc{Uniform} & 77.1 & 57.1 & 68.6 & 73.6 & 72.0 \\ 
      \textsc{Ind. Output} & 78.1 & 52.6 & 71.0 & 76.4 & \textbf{78.6}  \\
      \textsc{Pred. Conf.}  & 79.3 & 61.1 & 69.2 & 78.7 & \textbf{78.6} \\
      \midrule
      \parbox{2.5cm}{\textsc{Full Attention}\\(Ind. Output + Pred. Conf.)} & \textbf{82.7} & \textbf{66.7} & \textbf{77.4} & \textbf{89.1} & 66.7 \\
      \bottomrule 
    \end{tabular}
  \end{minipage} 
  \begin{minipage}{.55\linewidth}
    \vspace{0.18in}
    \begin{flushright}
    \includegraphics[width=0.9\textwidth]{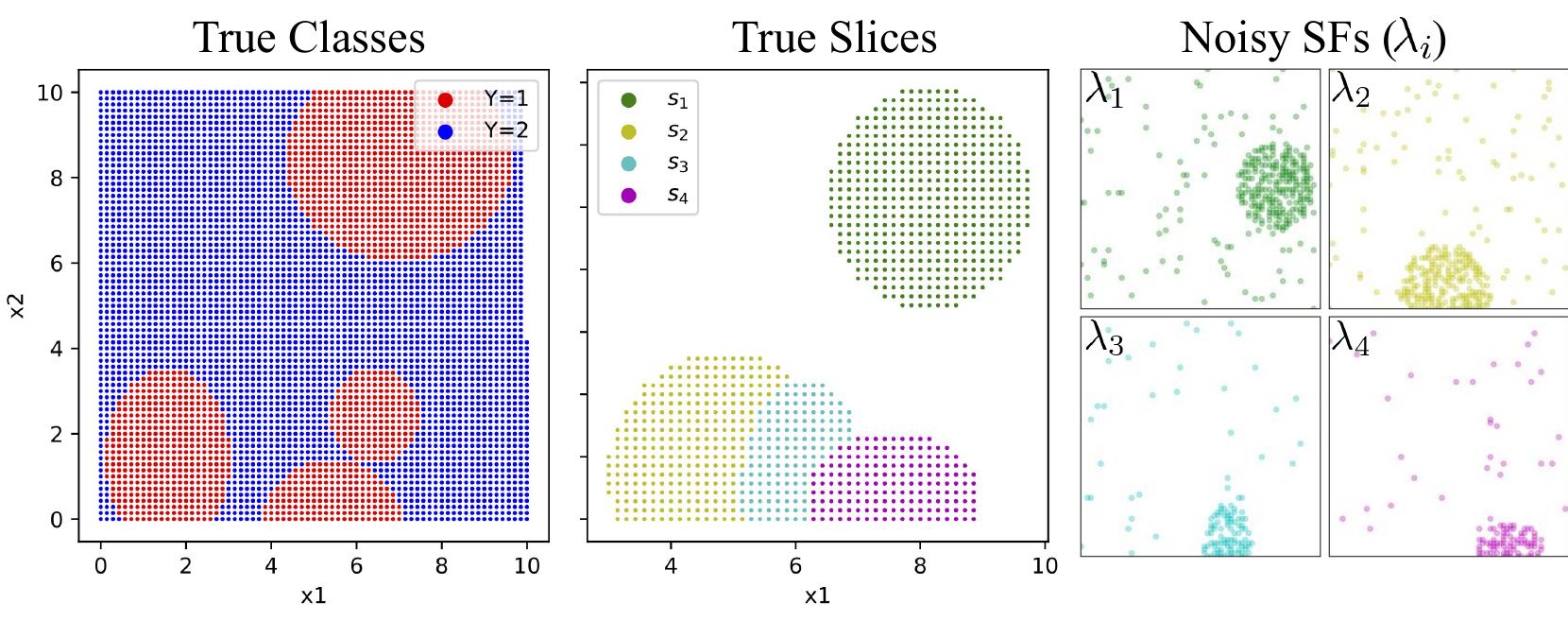}
    \end{flushright}
  \end{minipage}
\vspace{0.1in}
\captionof{figure}{
    \textbf{Architecture Ablation}:
    Using a synthetic, two-class dataset (Figure, left) with four randomly
    specified (size, shape, location) slices
    (Figure, middle), we specify corresponding, noisy SFs (Figure, right) and
    ablate specific model components by modifying the reweighting mechanism for
    slice expert representations. We compare overall/slice performance
    for uniform, indicator output, prediction confidence
    weighting, and the proposed attention weighting using all components.
    Our \textsc{Full Attention} approach performs most consistently on slices without worsening overall performance.
  }
    \label{fig:ablation_model}
\end{table}

\subsection{Model Architecture}

The \systemx architecture has six components. The key intuition is that we
will train a standard prediction model, which we call the {\em base task}. 
We then learn a representation for each slice that explains
how its predictions should differ from the representation of the base task---i.e., a \textit{residual}. An attention mechanism then combines these
representations to make a slice-aware prediction.

With this intuition in mind, the six components
(Figure~\ref{fig:architecture}) are: (a) a \textbf{backbone}, (b) a set of
$k$ \textbf{slice-indicator heads}, and (c) $k$ corresponding
\textbf{slice expert representations}, (d) a \textbf{shared slice prediction
head}, (e) a combined, \textbf{slice-aware representation}, and (f) a \textbf{prediction head}.
Each SRAM operates over any backbone architecture and represents a path through components (b) through (e).
We describe the architecture assuming a binary classification task (output dim. $c=1$):

\paragraph*{(a) Backbone:}
Our approach is agnostic to the neural network architecture, which we call the
\textit{backbone}, denoted $f_{\hat{w}}$, which is used primarily for feature extraction (e.g. the latest transformer for textual data, CNN for image data).
The backbone maps data points $x$ to a representation $z \in \mathbb{R}^d$.

\paragraph*{(b) Slice indicator heads:}
For each slice, an indicator head will output an input's slice membership. The model will later use this to reweight the
``expert'' slice representations based on the likelihood that an example is
in the corresponding slice. Each indicator head maps the backbone
representation, $z$, to a logit indicating slice-membership:
$\{q_i\}_{i=1,\dots,k} \in \mathbb{R}$.
Each slice indicator head is supervised by the output of a corresponding SF, $\lambda_i$.
For each example, we minimize the multi-label binary cross entropy loss
($\mathcal{L}_\textrm{CE}$) between the unnormalized logit output of each $q_i$
and $\lambda_i$: 
$\ell_{\text{ind}} = \sum_{i}^{k} \mathcal{L}_\textrm{CE} (q_i, \lambda_i)$
 
\paragraph*{(c) Slice expert representations:}
Each slice representation, $\{r_i\}_{i=1,\dots,k}$, will be treated as an
``expert'' feature for a given slice. We learn a linear mapping from the
backbone, $z$, to each $r_i \in \mathbb{R}^h$, where $d'$ is the size of all
slice expert representations.

\paragraph*{(d) Shared slice prediction head:}
A shared, slice prediction head, $g(\cdot)$, maps each slice expert
representation, $r_i$, to a logit, $\{p_i\}_{i=1,\dots,k}$, in the output space of
the base task:  $g(r_i) = p_i \in \mathbb{R}^{c}$, where $c=1$ for binary
classification. We train slice ``expert'' tasks using \textit{only}
examples belonging to the corresponding slice, as specified by $\lambda_i$. 
Because parameters in $g(\cdot)$ are shared, each representation, $r_i$, is
forced to \textit{specialize to the examples belonging to the slice}. We use the base task's ground truth label, $y$, to train this head with binary cross entropy loss:
$\ell_{\text{pred}} = \sum_i^k \lambda_i \mathcal{L}_\textrm{CE}(p_i, y)$

\paragraph*{(e) Slice-aware representation:}
For each example, the slice-aware representation is the combination of several ``expert'' slice
representations according to 1) the likelihood that the input is in the slice and 2) the confidence of the slice ``expert's'' prediction.
To explicitly model the residual from slice representations to the base
representation, we initialize a trivial ``base slice'' which consists of
\textit{all examples} so that we have the corresponding indicators, 
$q_\text{BASE}$, and predictors, $p_\text{BASE}$. 

Let $Q= \{q_1, \ldots, q_k, q_\text{BASE}\} \in \mathbb{R}^{k+1}$ 
be the vector of concatenated slice indicator logits,  
$P =\{p_1, \ldots, p_k, p_\text{BASE}\} \in \mathbb{R}^ {c \times k+1}$
be the vector of concatenated slice prediction logits, and 
$R = \{r_1,\dots,r_k,r_{\text{BASE}}\} \in \mathbb{R}^{h \times k + 1}$
be the $k+1$ stacked slice expert representations.
We compute our attention by combining the likelihood of slice membership, $Q$,
and the slice prediction confidence, which we interpret as a function of the logits---in the binary case $c=1$, we use $abs(P)$ as this confidence.
We then apply a Softmax to create soft attention
weights over the $k+1$ slice expert representations:
$a \in \mathbb{R}^{k+1} = \text{Softmax}(Q + abs(P))$.
Using a weighted sum, we then compute the combined, slice-aware representation:
$z' \in \mathbb{R}^{d'} = Ra$.

\paragraph*{(f) Prediction head}
Finally, we use our slice-aware representation $z'$ as the input to a final
linear layer, $h(\cdot)$, which we term the \textit{prediction head}, to make a prediction on the original, base task.
During inference, this prediction head makes the final prediction.
To train the prediction head, we minimize the cross entropy between the prediction head's
output, $h(z')$, and the base task's ground truth labels, $y$:
$\ell_\text{base} = \mathcal{L}_\textrm{CE}(h(z'), y)$.

Overall, the model is trained using loss values from all task heads:
$\ell_\text{train} = \ell_\text{base} + \ell_\text{ind} + \ell_\text{pred}$.
In Figure~\ref{fig:ablation_model}, we show ablations of this architecture in
a synthetic experiment varying the components in the reweighting mechanism---specifically, our described attention approach
outperforms using \textit{only} indicator outputs, \textit{only} predictor confidences, or uniform weights to reweight the slice representations.

\begin{table}[t]\
  \begin{minipage}{.36\linewidth}
  \small
  \begin{tabular}{lccc}\\
    \toprule
    \multirow{2}{*}{\textbf{Method}} & \multicolumn{3}{c}{Performance (F1 score)} \\
    & Overall & $S_1$ & $S_2$ \\
    \midrule
    \textsc{Vanilla} & 96.56 & 52.94 &  68.75 \\
    \textsc{DP}~\cite{ratner2016data} & 96.88 & 44.12 & 43.75 \\
    \textsc{HPS}~\cite{caruana1997multitask} & 96.72 & 50.00 & 75.00\\
    \textsc{MoE}~\cite{jacobs1991adaptive} & \textbf{98.48} & 88.24 & \textbf{87.50} \\
    \midrule
    \textsc{SBL} & 97.92 & \textbf{91.18} & 81.25 \\
    \bottomrule
\end{tabular}
\end{minipage}
\begin{minipage}{.60\linewidth}
\vspace{0.18in}
\begin{flushright}
\includegraphics[width=0.9\textwidth]{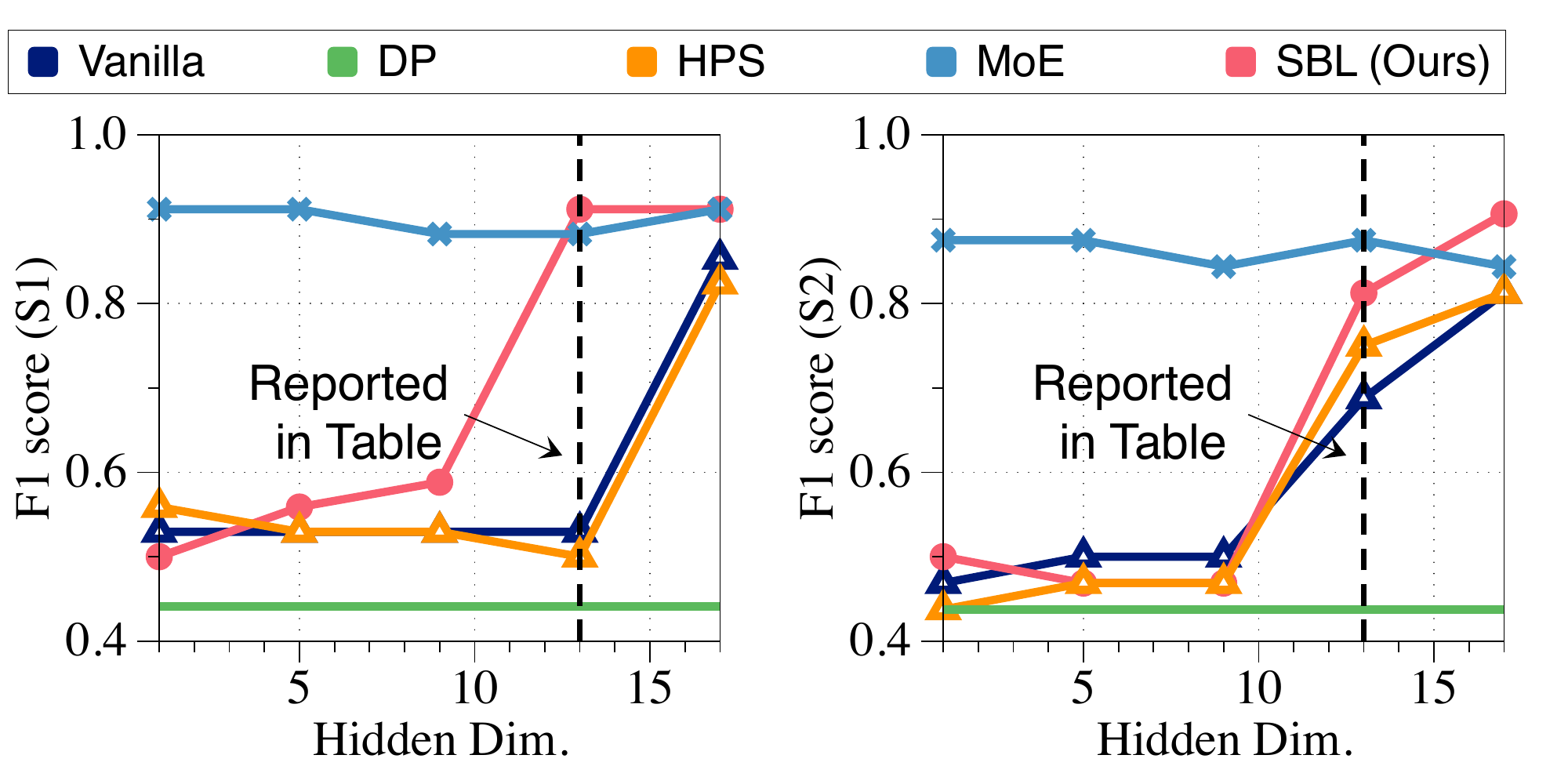}
\end{flushright}
\end{minipage}
\vspace{0.1in}

\captionof{figure}{ \textbf{Scaling with hidden feature representation dimensions.}
   We plot model quality versus the hidden
  dimension size. The slice-aware model
  (\textsc{SBL}) improves over \textit{hard parameter sharing} (\textsc{HPS})
  on both slices at a fixed hidden dimension size, while being close
  to \textit{mixture of experts} (\textsc{MoE}). Note: \textsc{MoE} has
  significantly
  more parameters overall, as it copies the entire model.
  } \label{fig:synthetics_results}
\vspace{-0.1in}

\end{table}

\begin{figure}[t]
\begin{center}
  \includegraphics[width=1.0\textwidth]{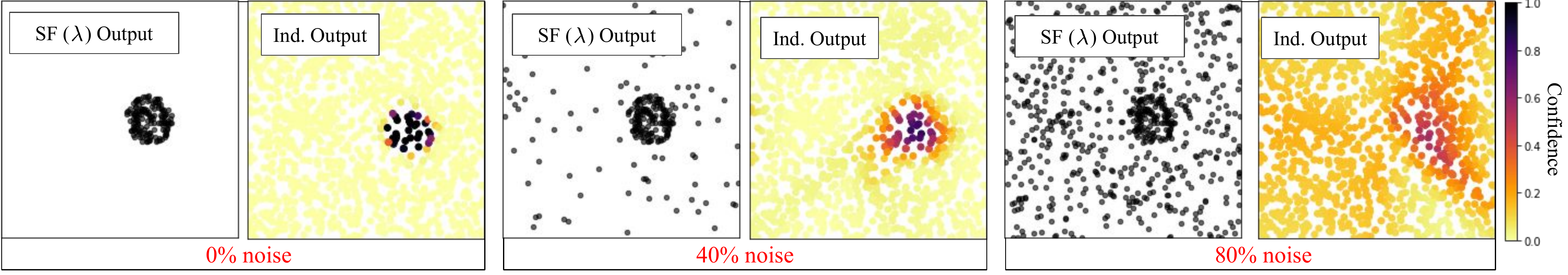}
  \caption{
    \textbf{Coping with Noise}:
    We test the robustness of our approach on a simple synthetic example.
    In each panel, we show noisy SFs (left) as binary points and the
    corresponding slice indicator's output (right) as a heatmap of
    probabilities. We show that the indicator assigns low relative
    probabilities on noisy (40\%, middle) samples and ignores a very noisy
    (80\%, right) SF, assigning relatively uniform scores to all samples.
  }
  \label{fig:noise_ablation}
\end{center}
 \vspace{-0.1in}
\end{figure}

\subsection{Synthetic data experiments}
\label{subsec:synthetics}

To understand the properties of \systemx (\textsc{SBL}), we validate our model and
its components (Figure~\ref{fig:architecture}) on a set of synthetic
data. In the results demonstrated in
Figure~\ref{fig:synthetics_overview}, we construct a dataset
$\mathcal{X} \in \mathbb{R}^2$ with a 2-way classification problem in
which over 95\% of the data are linearly separable.  We introduce two
minor perturbations along the decision boundary, which we define as
critical slices, $s_1$ and $s_2$.  Intuitively, examples that
fall within these slices follow different distributions
($P(\mathcal{Y} | \mathcal{X}, s_i)$) relative to
the overall data ($P(\mathcal{Y}|\mathcal{X})$).  For all models, the
shared backbone is defined as a 2-layer MLP architecture with a
backbone representation size $d=13$ and a final $ReLU$ non-linearity.
In \textsc{SBL}, the slice-expert representation is initialized with the same size: $d'=13$.

\paragraph{The model learns the slice-conditional label distribution 
$P(Y | s_i, X)$ from noisy SF inputs.}
We show in Figure~\ref{fig:synthetics_overview}b that the slices at the
perturbed decision boundary cannot be
learned in the general case, by a \textsc{Vanilla} model. As a result, we define two SFs, $\lambda_1$ and $\lambda_2$, to
target the slices of interest.
Because our attention-based model (\textsc{SBL}) is slice-aware,
it outperforms \textsc{Vanilla}, which has no notion of slices (Figure~\ref{fig:synthetics_overview}d).
Intuitively, if the model knows ``where'' in the 2-dim data space an example lives
(as defined by SFs), it can condition on slice-specific features as it makes a final, slice-aware prediction.
In Figure~\ref{fig:noise_ablation}, we observe our model's ability to cope with
noisy SF inputs: the indicator is robust to moderate amounts
of noise by ignoring noisy examples (middle); with extremely noisy inputs, it
disregards poorly-defined SFs by assigning relatively uniform weights (right).

\paragraph{Overall model performance does not degrade.}
The primary goal of the slice-aware model is to improve slice-specific
performance without degrading the model's existing capabilities. We
show that \textsc{SBL}
improves the overall score by \num{1.36} F1 points by
learning the proportionally smaller perturbations in the decision
boundary in addition to the more general linear boundary (Figure~\ref{fig:synthetics_results}, left). Further, we
note that we do not regress performance on individual slices.

\paragraph{Learning slice weights with features $P(Y|s_i,X)$ improves over doing so with only supervision source information $P(Y|s_i)$.}
A core assumption of our approach asserts that if the model learns improved
slice-conditional weights via $\lambda_i$, downstream slice-specific
performance will improve.
Data programming (DP) \cite{ratner2016data} is a popular weak supervision
approach deployed at numerous Fortune 500 companies \cite{bach2018snorkel,re2019overton}, in
which the weights of heuristics are learned solely from labeling source
information.
We emphasize that our setting provides the model with strictly more
information---in the data's feature representations---to learn such weights; we
show in Figure~\ref{fig:synthetics_results} (right) that increasing
representation size allows us to significantly outperform DP.

\paragraph{Attention weights learn from noisy $\lambda_i$ to combine slice
residual representations.}
\textsc{SBL} achieves improvements over methods that do
not aggregate slice information, as defined by each noisy $\lambda_i$.
Both the indicator outputs ($Q$) and prediction confidence ($abs(P)$) are
robustly combined in the attention mechanism.
Even a noisy indicator will be upweighted if the predictions are high
confidence, and if the indicator has high signal, even a slice expert making
poor predictions can benefit from underlying slice-specific features.
We show in Figure~\ref{fig:synthetics_results} that our method improves over
\textsc{HPS}, which is slice-aware, but has no way of combining slice
information despite increasingly noisy $\lambda_i$.
In contrast, our attention-based architecture is able to combine slice expert representations, as \textsc{SBL} sees improvements over \textsc{Vanilla} by \num{38.2}
slice-level F1 averaged across $s_1$ and $s_2$.

\paragraph{\textsc{SBL} demonstrates similar expressivity to MoE with much less
cost.}
With approximately half as many parameters, SBL comes within \num{6.25} slice-level F1 averaged across $s_1$ and $s_2$ of
\textsc{MoE}
(Figure~\ref{fig:synthetics_results}). With large backbone architectures, characterized by
$M$ parameters, and a large number of slices, $k$, \textsc{MoE} requires a
quadratically large number of parameters, because we initialize an entire
backbone for each slice. In contrast, all other models scale linearly
in parameters with $M$.

\section{Experiments}
\label{sec:experiments}
Compared to baselines using the same backbone architecture, we demonstrate that our approach successfully models slice
importance and improves slice-level performance without impacting
overall model performance. Then, we demonstrate our method's advantages in
aggregating noisy heuristics, compared to existing weak supervision literature.
We perform all empirical experiments on Google's Cloud infrastructure using NVIDIA V100 GPUs.

\begin{table}[t]
	\centering
  \tiny
	\begin{tabular}{lcccccccccccc}\\
		\toprule
		\multirow{2}{*}{\vspace*{8pt}\textbf{Dataset}}
      &
      \multicolumn{4}{c}{\textbf{\textsc{CoLA}} (Matthews Corr.~\cite{matthews1975comparison})} &
      \multicolumn{4}{c}{\textbf{\textsc{RTE}} (F1 Score)} &
      \multicolumn{4}{c}{\textbf{\textsc{CyDet}} (F1 Score)} \\
      \midrule
      &
      Param & \multirow{2}{*}{Overall (std)} & \multicolumn{2}{c}{{Slice Lift}} &
      Param & \multirow{2}{*}{Overall (std)} & \multicolumn{2}{c}{{Slice Lift }} &
      Param & \multirow{2}{*}{Overall (std)} & \multicolumn{2}{c}{{Slice Lift}} \\
      &
      Inc. & & Max & Avg &
      Inc. & & Max & Avg &
      Inc. & & Max & Avg \\
    \midrule
    \textsc{Vanilla} & -- & 57.8 ($\pm$1.3) & -- & -- & -- & 67.0 ($\pm$1.6) & --        & --      & -- & 39.4 ($\pm$-5.4)        & --     & -- \\
	  \textsc{HPS}~\cite{caruana1997multitask} & \textbf{12\%} & 57.4 ($\pm$2.1) & +12.7    & 1.1   & \textbf{10\%} & 67.9 ($\pm$1.8)   & \textbf{+12.7}    & +2.9   & \textbf{10\%} & 37.4 ($\pm$3.6) & +6.3  & -0.7 \\
		\textsc{Manual}  & \textbf{12\%} & 57.9 ($\pm$1.2) & +6.3     & +0.4   & \textbf{10\%} & 69.4 ($\pm$1.8)   & +10.7     & +4.2   & \textbf{10\%} & 36.9 ($\pm$4.2) & +6.3  & -1.7 \\
		\textsc{MoE}~\cite{jacobs1991adaptive} & $100\%$ & 57.2 ($\pm$0.9) & \textbf{+20.0}    & +1.3   &  100\% & 69.2 ($\pm$1.5)   & +10.9    & +3.9  & 100\% & \textrm{OOM}            & \textrm{OOM}     & \textrm{OOM} \\
		\midrule
		\textsc{SBL} & \textbf{12\%} & \textbf{58.3} ($\pm$0.7) & +19.0 & +\textbf{2.5} & \textbf{10\%} &\textbf{69.5} ($\pm$0.8) & +10.9 & \textbf{+4.6} & \textbf{10\%} & \textbf{40.9} ($\pm$3.9) & \textbf{+15.6} & \textbf{+2.3} \\
    \bottomrule
  \end{tabular}
  \vspace{0.1in}
  \caption{
    \textbf{Application Datasets}:
    We compare our model to baselines averaged over 5 runs with different seeds in natural language understanding and computer vision applications and note the relative increase in number of params for each method.
    We report the overall score and maximum relative improvement (denoted \textit{Lift}) over the \textsc{Vanilla} model for each of the slice-aware baselines.
    For some trials of \textsc{MoE}, our system ran out of GPU memory (denoted \textrm{OOM}).
  }
  \label{tab:main_results}
  \vspace{-0.1in}
\end{table}

\subsection{Applications}

Using natural language understanding (NLU) and computer vision (CV) datasets, we
compare our method to baselines commonly used in practice or the literature
to address slice-specific performance. 

\subsubsection{Baselines}
\label{subsubsec:baselines}

For each baseline, we first train the backbone parameters with a standard hyperparameter search over learning rate and $\ell_2$ regularization values.
Then, each method is initialized from the backbone weights and fine-tuned for a fixed number of epochs and the optimal hyperparameters.

\textsc{Vanilla}: A vanilla neural network
backbone is trained with a final prediction head to make predictions.
This baseline represents the de-facto approach used in deep learning modeling
tasks; it is unaware of slices information and neglects to
model them as a result. \\
\textsc{MoE}: We train a \textit{mixture of experts}
~\cite{jacobs1991adaptive}, where each \textit{expert} is a separate
\textsc{Vanilla} model trained on a data subset specified by the SF, $\lambda_i$. A gating
network~\cite{sigaud2015gated} is then trained to combine expert predictions
into a final prediction. \\
\textsc{HPS}: In the style
of multi-task learning, we model slices as separate task heads
with a shared backbone trained via \textit{hard parameter sharing}. Each slice task
performs the same prediction task, but they are trained on subsets of data
corresponding to $\lambda_i$. In this approach, backpropagation from different
slice tasks is intended to encourage a slice-aware representation bias
\cite{ruder2017overview,caruana1997multitask}. \\
\textsc{Manual}: To simulate the manual effort required to
tune slice-specific hyperparameters, we leverage the same architecture as \textsc{HPS}
and grid search over loss term multipliers, $\alpha \in \{2, 20, 50, 100\}$,
for underperforming slices based on \textsc{Vanilla} model predictions (i.e. $\text{score}_{\text{overall}} -
\text{score}_{\text{slice}} \ge 5$ F1).

\subsubsection{Datasets}
\textbf{NLU Datasets.} We select slices based
on independently-conducted error analyses~\cite{kim2019probing} (Appendix~\ref{apx:sfs}).
In \textbf{Corpus of Linguistic Acceptability (\textsc{CoLA})}
~\cite{warstadt2018neural}, the task is to predict whether a
sentence is linguistically acceptable (i.e. grammatically); we measure
performance using the Matthews correlation coefficient
~\cite{matthews1975comparison}.  Natural slices might occur as questions or
long sentences, as corresponding examples might consist of non-standard or
challenging sentence structure. Since ground truth
test labels are not available for this task (they are held out in evaluation
servers~\cite{wang2018glue}), we sample to create
data splits with \num{7.2}K/\num{1.3}K/\num{1}K train/valid/test sentences,
respectively. To properly evaluate slices of interest, we ensure that the
proportions of examples in ground truth slices are consistent across splits.
In \textbf{Recognizing Textual Entailment (\textsc{RTE})}~\cite{wang2018glue,
dagan2005pascal,bar2006second,giampiccolo2007third,bentivogli2009fifth}, the
task is to predict whether or not a premise sentence entails a
hypothesis sentence. Similar to \textsc{CoLA}, we create
our own data splits and use \num{2.25}K/\num{0.25}K/\num{0.275}K
train/valid/test sentences, respectively. 
Finally, in a user study where we work with practitioners tackling the \textbf{SuperGlue}~\cite{wang2019superglue} benchmark, we leverage
\systemx to improve state-of-the-art model quality on benchmark submissions.

\textbf{CV Dataset.} In the image domain, we evaluate on an autonomous
vehicle dataset called \textbf{Cyclist Detection for Autonomous Vehicles
(\textsc{CyDet})} \cite{masalov2018cydet}. We leverage clips in a
self-driving video dataset to detect whether a cyclist (person plus
bicycle) is present at each frame. We select one independent clip for
evaluation, and the remainder for training; for valid/test splits, we
select alternating batches of five frames each from the evaluation clip. 
We preprocess the dataset with an open-source implementation of
Mask R-CNN~\cite{massa2018mrcnn} to provide
metadata (e.g. presence of traffic lights, benches), which serve as slice
indicators for each frame.

\subsubsection{Results}

\textbf{Slice-aware models improve slice-specific performance.}
We see in Table~\ref{tab:main_results} that each slice-aware model (\textsc{HPS}, \textsc{Manual}, \textsc{MoE}, \textsc{SBL}) largely improves over the naive model.

\textbf{\textsc{SBL} improves overall performance.} We also observe that \textsc{SBL} improves overall performance for each of the datasets.
This is likely because the chosen slices were explicitly modeled from error analysis papers, and explicitly modeling ``error'' slices led to improved
overall performance.

\textbf{\textsc{SBL} learns slice expert representations consistently.}
While \textsc{HPS} and \textsc{Manual} perform well on some slices, they
exhibit much higher variance compared to \textsc{SBL} and \textsc{MoE} 
(as denoted by the std. in Table~\ref{tab:main_results}).
These baselines lack an attention mechanism to reweight slice representations in a consistent way; instead, they rely purely on representation bias from slice-specific heads to improve slice-level
performance. Because these representations are not modeled explicitly, 
improvements are largely driven by chance, and this approach risks worsening
performance on other slices or overall.

\textbf{\textsc{SBL} improves performance with a parameter-efficient representation.}
For \textbf{CoLA} and \textbf{RTE} experiments, we used the \texttt{BERT-base}
~\cite{devlin2018bert} architecture with 110M parameters; for \textbf{CyDet},
we used \texttt{ResNet-18}~\cite{he2016deep}.
For each additional slice, \textsc{SBL} requires a 7\% and 5\% increase in relative parameter count in the BERT and ResNet architectures, respectively (total relative parameter increase reported in Table~\ref{tab:main_results}).
As a comparison, \textsc{HPS} requires the same relative increase in
parameters per slice.
\textsc{MoE} on the other hand, increases relative number of parameters by
100\% per slice for both architectures.
With limited increase in model size, \textsc{SBL} outperforms or
matches all other baselines, including \textsc{MoE}, which requires an order of
magnitude more parameters.

\textbf{\textsc{SBL} improves state-of-the-art quality models with 
slice-aware representations.} In a submission to SuperGLUE evaluation servers, we leverage the same \texttt{BERT-large} architecture as previous submissions and observe improvements on NLU tasks: +\num{3.8/+2.8} avg. F1/acc.
on CB~\cite{de2019commitmentbank}, +\num {2.4} acc. on COPA~\cite{roemmele2011choice}, +\num{2.5} acc. on WiC~\cite{pilehvar2018wic}, and +\num{2.7} on the aggregate benchmark score.

\subsection{Weak Supervision Comparisons}
To contextualize our contributions in the weak supervision literature,
we compare directly to Data Programming (\textsc{DP})~\cite{ratner2018snorkel},
a popular approach for reweighting user-specified heuristics using supervision
source information~\cite{ratner2016data}. 
We consider two text-based relation extraction datasets: 
\textbf{Chemical-Disease Relations (CDR)},\cite{wei2015overview}, 
in which we identify causal links between chemical and disease entities in a
dataset of PubMed abstracts, and \textbf{Spouses} \cite{corney2016million}, in
which we identify mentions of spousal relationships using preprocessed pairs of
person mentions from news articles (via Spacy~\cite{honnibal-johnson:2015:EMNLP}).
In both datasets, we leverage the exact noisy linguistic patterns and distant
supervision heuristics provided in the open-source implementation of
\textsc{DP}. Rather than voting on
a particular class, we repurpose the provided labeling functions as binary
slice indicators for our model. We then train our slice-aware model on the
probabilistic labels aggregated from these heuristics.

\textbf{\textsc{SBL} improves over current weak supervision methods.}
Treating the noisy heuristics as slicing functions, we observe
lifts of up to \num{1.3 F1} overall and \num{15.9 F1} on heuristically-defined
slices. We reproduce the \textsc{DP}~\cite{ratner2018snorkel} setup to obtain overall scores of F1=\num{$41.9$} on
\textbf{Spouses} and F1=\num{$56.4$} on \textbf{CDR}.
Using \systemx, we improve to \num{$42.8$ ($+0.9$) and $57.7$ ($+1.3$) F1},
respectively. Intuitively, we can explain this improvement, because 
\textsc{SBL} has access to features of the data belonging to slices whereas
\textsc{DP} relies only on the source information of each heuristic.

\section{Conclusion}
\label{sec:conclusion}
We introduced the challenge of improving slice-specific performance
without damaging the overall model quality, and proposed the first programming
abstraction and machine learning model to support these actions. We
demonstrated that the model could be used to push the state-of-the-art quality.
In our analysis, we can explain consistent gains in the \systemx paradigm, as
our attention mechanism has access to a rich set of deep features,
whereas existing weak supervision paradigms have no way to access this information.
We view this work in the context of programming models that sit on top of
traditional modeling approaches in machine learning systems.

\small{
  \textit{Acknowledgements} 
  We would like to thank Braden Hancock, Feng Niu, and Charles Srisuwananukorn for many helpful discussions, tests, and collaborations throughout the development of slicing.
We gratefully acknowledge the support of DARPA under Nos. FA87501720095 (D3M), FA86501827865 (SDH), FA86501827882 (ASED), NIH under No. U54EB020405 (Mobilize), NSF under Nos. CCF1763315 (Beyond Sparsity) and CCF1563078 (Volume to Velocity), ONR under No. N000141712266 (Unifying Weak Supervision), the Moore Foundation, NXP, Xilinx, LETI-CEA, Intel, Microsoft, NEC, Toshiba, TSMC, ARM, Hitachi, BASF, Accenture, Ericsson, Qualcomm, Analog Devices, the Okawa Foundation, and American Family Insurance, Google Cloud, Swiss Re, and members of the Stanford DAWN project: Teradata, Facebook, Google, Ant Financial, NEC, SAP, VMWare, and Infosys. The U.S. Government is authorized to reproduce and distribute reprints for Governmental purposes notwithstanding any copyright notation thereon. Any opinions, findings, and conclusions or recommendations expressed in this material are those of the authors and do not necessarily reflect the views, policies, or endorsements, either expressed or implied, of DARPA, NIH, ONR, or the U.S. Government.
} \label{sec:acks}

\small{
  \bibliographystyle{plain}
  \bibliography{main}
}

\newpage
\renewcommand{\thesection}{A\arabic{section}}
\setcounter{section}{0} %
\section{Appendix}
\label{sec:appendix}
\subsection{Model Characteristics}
\label{apdx:modelchars}
We include summarized model characteristics and the associated baselines to supplement Sections~\ref{subsec:synthetics} and \ref{subsubsec:baselines}.
\begin{table}[h]
  \centering
  \begin{tabular}{lccccc}\\
    \toprule
    \multirow{2}{*}{\vspace*{8pt}
      \textbf{Method}}&
      Slice-aware &
      \parbox{1.6cm}{No manual tuning} &
      \parbox{1.5cm}{Weighted slice info.} &
      \parbox{2.8cm}{Avoids copies of model ($M$ params)} &
      Num. Params  \\
    \midrule
    \textsc{Vanilla} &  &  \checkmark &  & \checkmark & $O(M+r)$ \\
    \textsc{HPS} & \checkmark & \checkmark &  & \checkmark & $O(M+kr)$ \\
    \textsc{Manual} & \checkmark &  & \checkmark & \checkmark & $O(M+kr)$ \\
    \textsc{MoE} & \checkmark & \checkmark & \checkmark & & $O(kM+kr)$  \\
    \textsc{SBL} & \checkmark & \checkmark & \checkmark & \checkmark & $O(M+krd')$ \\
    \bottomrule
  \end{tabular}
  \vspace{0.1in}
  \caption{
    \textbf{Model characterizations}: We characterize each model's advantages/limitations and compute the number of parameters for each baseline model, given $k$ slices, $M$ backbone parameters, feature representation $z$ dimension $r$, and slice expert representation $p_i$ dimension $d'$.
  }
  \label{tab:model_params}
\end{table}

\subsection{Slicing Function (SF) Construction}
\label{apx:sfs}
We walk through specific examples of SFs written for a number of our applications.

\paragraph{Textual SFs} For text-based applications (\textsc{CoLA, RTE}), we write SFs over pairs of sentences for each task.
Following dataset convention, we denote the first sentence as the \textit{premise} and the second as the \textit{hypothesis} where appropriate.
Then, SFs are written, drawing largely from existing error analysis~\cite{kim2019probing}.
For instance, we might expect certain questions to be especially difficult to formulate in a language acceptability task.
So, we write the following SF to heuristically target \textit{where} questions:
\begin{python}
def SF_where_question(premise, hypothesis):
    # triggers if "where" appears in sentence
    sentences = premise + hypothesis
    return "where" in sentences.lower()
\end{python}

In some cases, we write SFs over both sentences at once.
For instance, to capture possible errors in article references (e.g. \textit{the Big Apple} vs \textit{a big apple}), we specify a slice where multiple instances of the same article appear in provided sentences:
\begin{python}
def SF_has_multiple_articles(premise, hypothesis):
    # triggers if a sentence has more than one occurrence of the same article
    sentences = premise + hypothesis
    multiple_a = sum([int(x == "a") for x in sentences.split()]) > 1
    multiple_an = sum([int(x == "an") for x in sentences.split()]) > 1
    multiple_the = sum([int(x == "the") for x in sentences.split()]) > 1
    return multiple_a or multiple_an or multiple_the
\end{python}

\paragraph{Image-based SFs} For computer vision applications, we leverage image metadata and bounding box attributes, generated from an off-the-shelf Mask R-CNN~\cite{massa2018mrcnn}, to target slices of interest.
\begin{python}
def SF_bus(image):
    # triggers if a "bus" appears in the predictions of the noisy detector
    outputs = noisy_detector(image)
    return "bus" in outputs
\end{python}

We note that these potentially expensive detectors are \textit{non-servable}---they run offline, and our model uses learned indicators at inference time.
Despite the detectors' noisy predictions, our model is able to to reweight representations appropriately.

\subsection{\textsc{CoLA} SFs}
\textsc{CoLA} is a language acceptability task based on linguistics and grammar for individual sentences.
We draw from error analysis which introduces several linguistically imortance slices for language acceptability via a series of challenge tasks.
Each task consists of synthetically generated examples to measure model evaluation on specific slices.
We heuristically define SFs to target subsets of data corresponding to each challenge, and include the full set of SFs derived from each category of challenge tasks:
\begin{itemize}
\item \textbf{Wh-words}: This task targets sentences containing \textit{who, what, where, when, why, how}.
We exclude \textit{why} and \textit{how} below because the \textsc{CoLA} dataset does not have enough examples for proper training and evaluation of these slices.
\begin{python}
def SF_where_in_sentence(sentence):
    return "where" in sentence

def SF_who_in_sentence(sentence):
    return "who" in sentence

def SF_what_in_sentence(sentence):
    return "what" in sentence

def SF_when_in_sentence(sentence):
    return "when" in sentence
\end{python}
\item \textbf{Definite-Indefinite Articles}: This challenge measures the model based on different combinations of definite (\textit{the}) and indefinite (\textit{a,an}) articles in a sentence (i.e. swapping definite for indefinite articles and vice versa).
We target containing multiple uses of a definite (\textit{the}) or indefinite article (\textit{a, an}):
\begin{python}
def SF_has_multiple_articles(sentence):
    # triggers if a sentence has more than one occurrence of the same article
    multiple_indefinite = sum([int(x == "a") for x in sentence.split()]) > 1 or sum([int(x == "an") for x in sentence.split()]) > 1
    multiple_definite = sum([int(x == "the") for x in sentence.split()]) > 1

    return multiple_indefinite or multiple_definite
\end{python}
\item \textbf{Coordinating Conjunctions}: This task seeks to measure correct usage of coordinating conjunctions (\textit{and, but, or}) in context.
We target the presence of these words in both sentences.
\begin{python}
def and_in_sentence(sentence):
    return "and" in sentence

def but_in_sentence(sentence):
    return "but" in sentence

def or_in_sentence(sentence):
    return "or" in sentence
\end{python}
\item \textbf{End-of-Sentence}: This challenge task measures a model's ability to identify coherent sentences or sentence chunks after removing puctuation.
We heuristically target this slice by identifying particularly short sentences and those that end with verbs and adverbs.
We use off-the-shelf parsers (i.e. Spacy~\cite{honnibal-johnson:2015:EMNLP}) to generate part-of-speech tags.
\begin{python}
def SF_short_sentence(sentence):
    # triggered if sentence has fewer than 5 tokens
    return len(sentence.split()) < 5

# Spacy tagger
def get_spacy_pos(sentence):
  import spacy
  nlp = spacy.load("en_core_web_sm")
  return nlp(sentence).pos_

def SF_ends_with_verb(sentence):
    # remove last token, which is always punctuation
    sentence = sentence[:-1]
    return get_spacy_pos(sentence)[-1] == "VERB"

def SF_ends_with_adverb(sentence):
    # remove last token, which is always punctuation
    sentence = sentence[:-1]
    return get_spacy_pos(sentence)[-1] == "ADVERB"
\end{python}
\end{itemize}

\subsection{\textsc{RTE} SFs}
Similar to \textsc{CoLA}, we use challenge tasks from NLI-based error analysis~\cite{kim2019probing} to write SFs over the textual entailment (\textsc{RTE}) dataset.
\begin{itemize}
\item \textbf{Prepositions}: In one challenge, the authors swap prepositions in the dataset with prepositions in a manually-curated list.
The list in its entirety spans a large proportion of the \textsc{RTE} dataset, which would constitute a very large slice. We find it more effective to separate these prepositions into \textit{temporal} and \textit{possessive} slices.
\begin{python}
def SF_has_temporal_preposition(premise, hypothesis):
    temporal_prepositions = ["after", "before", "past"]
    sentence = premise + sentence
    return any([p in sentence for p in temporal_prepositions])

def SF_has_possessive_preposition(premise, hypothesis):
    possessive_prepositions = ["inside of", "with", "within"]
    sentence = premise + sentence
    return any([p in sentence for p in possessive_prepositions])
\end{python}

\item \textbf{Comparatives}: One challenge chooses sentences with specific comparative words and mutates/negates them.
We directly target keywords identified in their approach.
\begin{python}
def SF_is_comparative(premise, hypothesis):
    comparative_words = ["more", "less", "better", "worse", "bigger", "smaller"]
    sentence = premise + hypothesis
    return any([p in sentence for p in comparative_words])
\end{python}

\item \textbf{Quantification}: One challenge tests natural language understanding with common quantifiers.
We target common quantifiers in both the combined premise/hypothesis and in \textit{only} the hypothesis.
\begin{python}
def is_quantification(premise, hypothesis):
    quantifiers = ["all", "some", "none"]
    sentence = premise + hypothesis
    return any([p in sentence for p in quantifiers])

def is_quantification_hypothesis(premise, hypothesis):
    quantifiers = ["all", "some", "none"]
    return any([p in hypothesis for p in quantifiers])
\end{python}

\item \textbf{Spatial Expressions}: This challenge identifies spatial relations between entities (i.e. \textit{A is to the left of B}).
We exclude this task from our slices, because such slices do not account for enough examples in the \textit{RTE} dataset.

\item \textbf{Negation}: This challenge task identifies whether natural language inference models can handle negations.
We heuristically target this slice via a list of common negation words from a top result in a web search.
\begin{python}
def SF_common_negation(premise, hypothesis):
    # Words from https://www.grammarly.com/blog/negatives/
    negation_words = [
        "no",
        "not",
        "none",
        "no one",
        "nobody",
        "nothing",
        "neither",
        "nowhere",
        "never",
        "hardly",
        "scarcely",
        "barely",
        "doesnt",
        "isnt",
        "wasnt",
        "shouldnt",
        "wouldnt",
        "couldnt",
        "wont",
        "cant",
        "dont",
    ]
    sentence = premise + hypothesis
    return any([x in negation_words for x in sentence])
\end{python}

\item \textbf{Premise/Hypothesis Length}: Finally, separate from the cited error analysis, we target different length hypotheses and premises as an additional set of slicing tasks.
In our own error analysis of the \textsc{RTE} model, we found these represented intuitive slices: long premises are typically harder to parse for key information, and shorter hypotheses tend to share syntactical structure.
\begin{python}
def SF_short_hypothesis(premise, hypothesis):
    return len(hypothesis.split()) < 5

def SF_long_hypothesis(premise, hypothesis):
    return len(hypothesis.split()) > 100

def SF_short_premise(premise, hypothesis):
    return len(premise.split()) < 15

def SF_long_premise(premise, hypothesis):
    return len(premise.split()) > 100
\end{python}

\end{itemize}

\subsection{\textsc{CyDet} SFs}
For the cyclist detection dataset, we identify subsets that correspond to other objects in the scene using a noisy detector (i.e. an off-the-shelf Mask R-CNN~\cite{massa2018mrcnn}).
\begin{python}
# define noisy detector
def noise_detector(image):
  probs = mask_rcnn.forward(image)

  # threshold predictions
  preds = []
  for object in classes:
      if probs["object"] > 0.5:
          preds.append(object)
  return preds

# Cyclist Detection SFs
def SF_bench(image):
    outputs = noisy_detector(image)
    return "bench" in outputs

def SF_truck(image):
    outputs = noisy_detector(image)
    return "truck" in outputs

def SF_car(image):
    outputs = noisy_detector(image)
    return "car" in outputs

def SF_bus(image):
    outputs = noisy_detector(image)
    return "bus" in outputs

def SF_person(image):
    outputs = noisy_detector(image)
    return "person" in outputs

def SF_traffic_light(image):
    outputs = noisy_detector(image)
    return "traffic light" in outputs

def SF_fire_hydrant(image):
    outputs = noisy_detector(image)
    return "fire hydrant" in outputs

def SF_stop_sign(image):
    outputs = noisy_detector(image)
    return "stop sign" in outputs

def SF_bicycle(image):
    outputs = noisy_detector(image)
    return "bicycle" in outputs
\end{python}

\begin{figure}[h]
  \centering
  \includegraphics[width=1.0\textwidth]{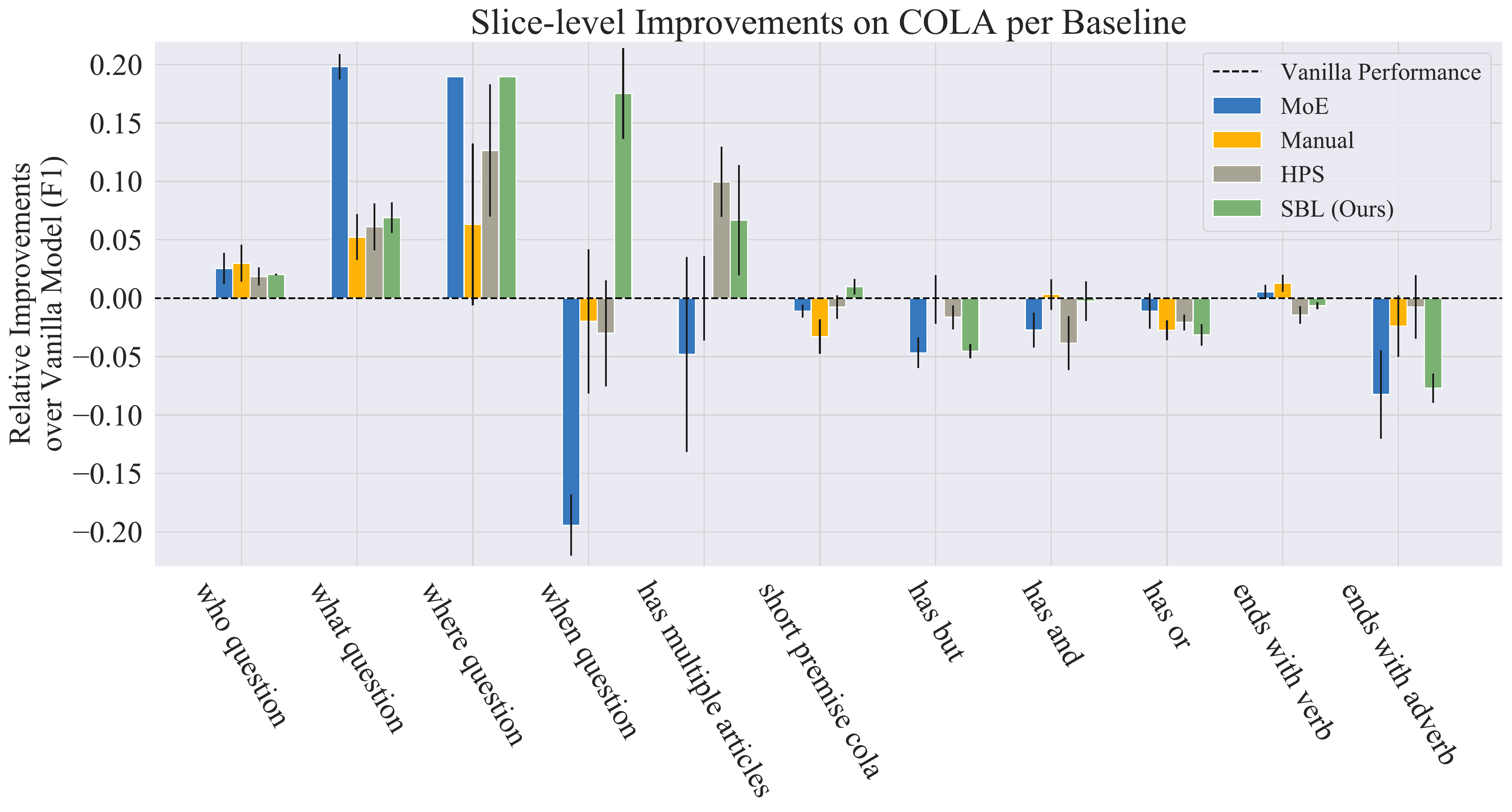}
  \includegraphics[width=1.0\textwidth]{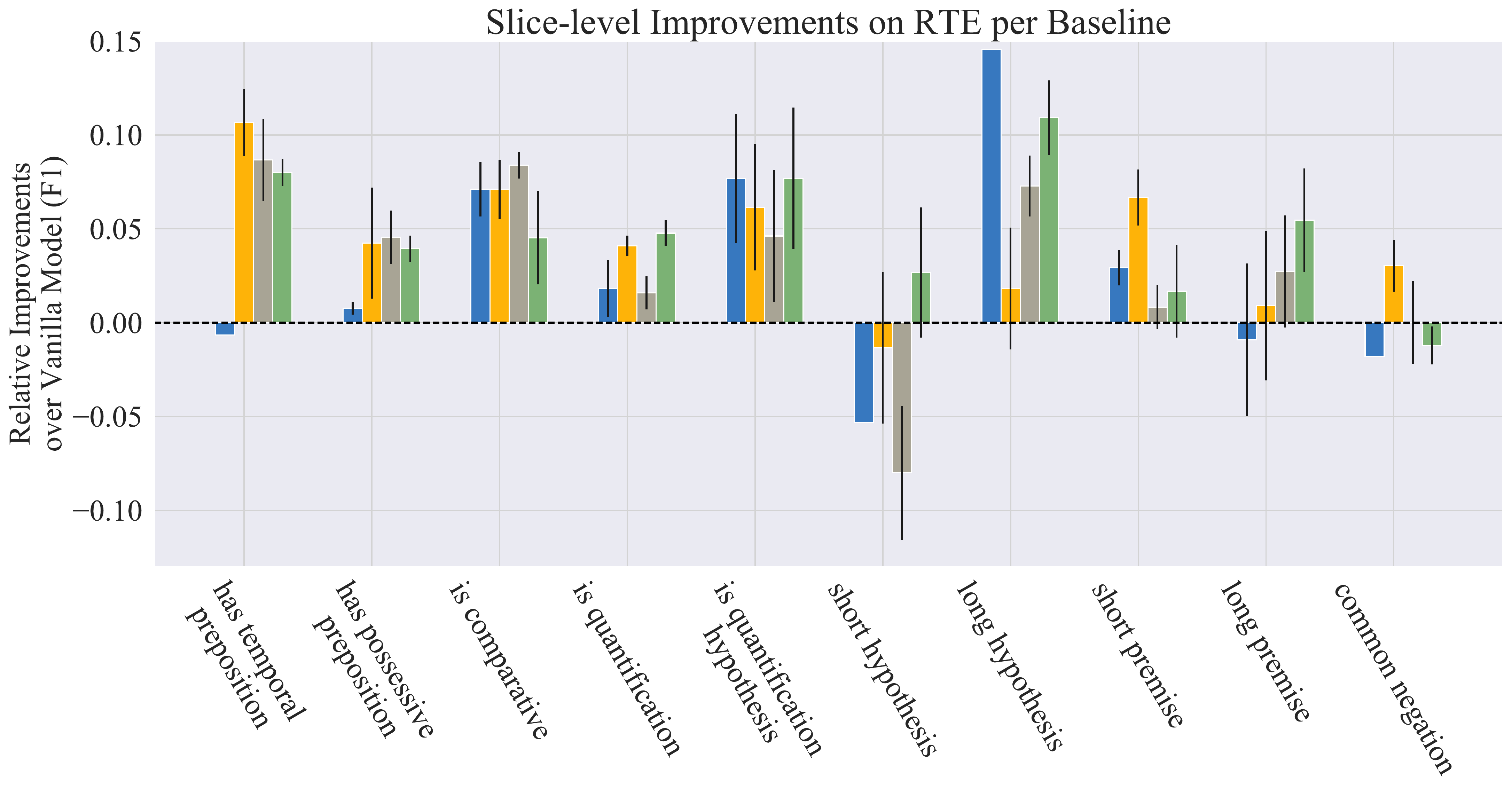}
  \includegraphics[width=1.0\textwidth]{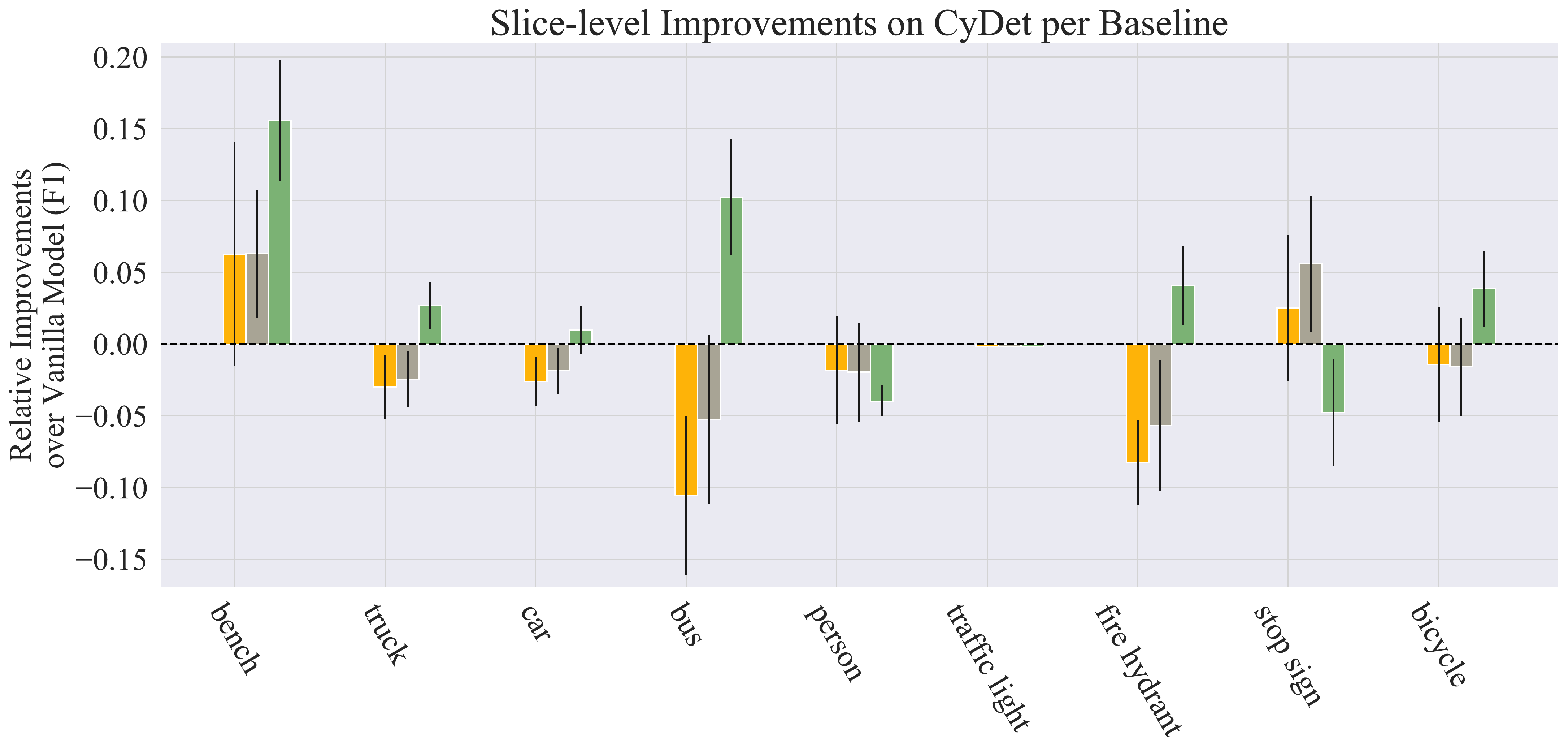}
  \caption{
    For each application dataset (Section 4.1) we report all relative, slice-level metrics compared to \textsc{Vanilla} for each model.
  }
  \label{fig:glue_slice_results}
\end{figure}
\subsection{Slice-specific Metrics}
We visualize slice-specific metrics across each application dataset, for each method of comparison.
We report the corresponding aggregate metrics in Figure 1 (below).

In \textsc{CoLA}, we see that \textsc{MoE} and \textsc{SBL} exhibit the largest slice-specific gains, and also overfit on the same slice \textit{ends with adverb}.
In \textsc{RTE}, we see that \textsc{SBL} improves performance on all slices except \textit{common negation}, where it falls less than a point below \textsc{Vanilla}.
On \textsc{CyDet}, we see the largest gains for \textsc{SBL} on \textit{bench} and \textit{bus} slices---in particular, we are able to improve in cases where the model might able to use the presence of these objects to make more informed decisions about whether a cyclist is present.
Note: because the \textsc{MoE} model on \textsc{CyDet} encounters an ``Out of Memory" error, the corresponding (blue) data bar is not available for this dataset.

\clearpage

\end{document}